\documentclass{article} % For LaTeX2e
\usepackage[utf8]{inputenc}
\usepackage{iclr2018_conference,times}
\usepackage{color}

\usepackage{amssymb}  % assumes amsmath package installed
\usepackage{hyperref}
\usepackage{ucs}
\usepackage{comment}
\usepackage{amsmath}
\usepackage{pbox}%fait des boites
\usepackage{tcolorbox}
\usepackage{caption}
\usepackage{capt-of}
\usepackage{bm}
\usepackage{graphicx}

\usepackage{array}
\usepackage{multirow}
\usepackage{colortbl}
\usepackage{tikz}
\usepackage{xspace}
\usepackage[caption=false,font=footnotesize]{subfig}

\usepackage{subfig}

\def\bbbe{{\rm I\!E}}

\newcommand{\Esp}{{\bbbe}{}}
\newcommand{\mymath}[1]{\ensuremath{#1}\xspace}
\newcommand{\mymathbf}[1]{\mymath{\mathbf{\boldsymbol{#1}}}}

\renewcommand{\vec}[1]{\mymathbf{#1}}

\newcommand{\vs}{\vec{s}}

\newcommand\wi[1]{$\circ$}
\newcommand\bu[1]{$\bullet$}
\newcommand\ot[1]{$\star$}
\newcommand\spa[1]{$\spadesuit$}
\newcommand\dn[1]{.}

\newcounter{cbox} 
\newcommand{\cbox}{\arabic{cbox}}

\newcounter{cmes} 
\newcommand{\cmes}{\arabic{cmes}}

\usepackage{algorithmic}
\usepackage{algorithm} 

\makeatletter
\newcounter{algorithmbis}
\renewcommand{\thealgorithmbis}{\thesection.\arabic{algorithmbis}}
\def\algorithmbis{\@ifnextchar[{\@algorithmbisa}{\@algorithmbisb}}
\def\@algorithmbisa[#1]{%
  \refstepcounter{algorithmbis}
  \trivlist
  \leftmargin\z@
  \itemindent\z@
  \labelsep\z@
  \item[\parbox{\linewidth}{%
    \hrule
    \hrule
    \noindent\strut\textbf{Algorithm \thealgorithmbis} #1
    \hrule
  }]\hfil\vskip0em%
}
\def\@algorithmbisb{\@algorithmbisa[]}

\makeatother

\definecolor{myred}{rgb}{0.8,0,0}
\definecolor{mygreen}{rgb}{0,0.6,0}
\definecolor{myblue}{rgb}{0,0,0.7}

\date{}

\usepackage{geometry}
\usepackage{tcolorbox}
\usepackage{tikz}
\usepackage{enumitem}
\usepackage{xcolor}

\newcommand{\vth}{\mymath{\vec \theta}}
\newcommand{\vx}{\mymath{\vec x}}
\newcommand{\vy}{\mymath{\vec y}}

\newcommand{\s}{\mymath{\vec s}}

\newcommand{\delth}{\mymath{\delta \vth}}

\newcommand{\hth}{\mymath{\vec h_{\vth}}}
\newcommand{\Dth}{\mymath{\Delta_{\vth}}}
\newcommand{\hdth}{\mymath{\vec h_{\vth + \delth}}}
\newcommand{\grad}{\mymath{\nabla_{\vth}}}

\newcounter{cbo} 
\newcommand{\cbo}{\arabic{cbo}}

\begin{document}
\title{First-order and second-order variants of the gradient descent in a unified framework}
\author{Thomas Pierrot*, Nicolas Perrin-Gilbert* and Olivier Sigaud\\
Sorbonne Universit\'e, CNRS UMR 7222,\\
      Institut des Syst\`emes Intelligents et de Robotique, F-75005 Paris, France\\
      {\tt thomas.pierrot@student.isae-supaero.fr}\\
      {\tt \{nicolas.perrin, olivier.sigaud\}@sorbonne-universite.fr}\\
      \textbf{*Equal contribution.}
    }
\maketitle
    \begin{abstract}
    In this paper, we provide an overview of first-order and second-order variants of the gradient descent method that are commonly used in machine learning.
    We propose a general framework in which 6 of these variants can be interpreted as different instances of the same approach. They are the vanilla gradient descent, the classical and generalized Gauss-Newton methods, the natural gradient descent method, the gradient covariance matrix approach, and Newton's method.
    Besides interpreting these methods within a single framework, we explain their specificities and show under which conditions some of them coincide. 
    \end{abstract}

\section{Introduction}
\label{sec:intro}

Machine learning generally amounts to solving an optimization problem where a loss function has to be minimized. As the problems tackled are getting more and more complex (nonlinear, nonconvex, etc.), fewer efficient algorithms exist, and the best recourse seems to rely on iterative schemes that exploit first-order or second-order derivatives of the loss function to get successive improvements and converge towards a local minimum. This explains why variants of gradient descent are becoming increasingly ubiquitous in machine learning and have been made widely available in the main deep learning libraries, being the tool of choice to optimize deep neural networks. Other types of local algorithms exist when no derivatives are known \citep{sigaud2018policy}, but in this paper we assume that some derivatives are available and only consider first order gradient-based or second order Hessian-based methods.

Among these methods, vanilla gradient descent strongly benefits from its computational efficiency as it simply computes partial derivatives at each step of an iterative process. Though it is widely used, it is limited for two main reasons: it depends on arbitrary parameterizations and may diverge or converge very slowly if the step size is not properly tuned.
To address these issues, several lines of improvement exist. Here, we focus on two of them. On the one hand, 
first-order methods such as the natural gradient introduce particular metrics to restrict gradient steps and make them independent from parametrization choices \citep{amari98natural}. On the other hand, second-order methods use the Hessian matrix of the loss or its approximations to take into account its local curvature.

\begin{figure}[!ht]
	\centering
	\includegraphics[width=0.67\textwidth]{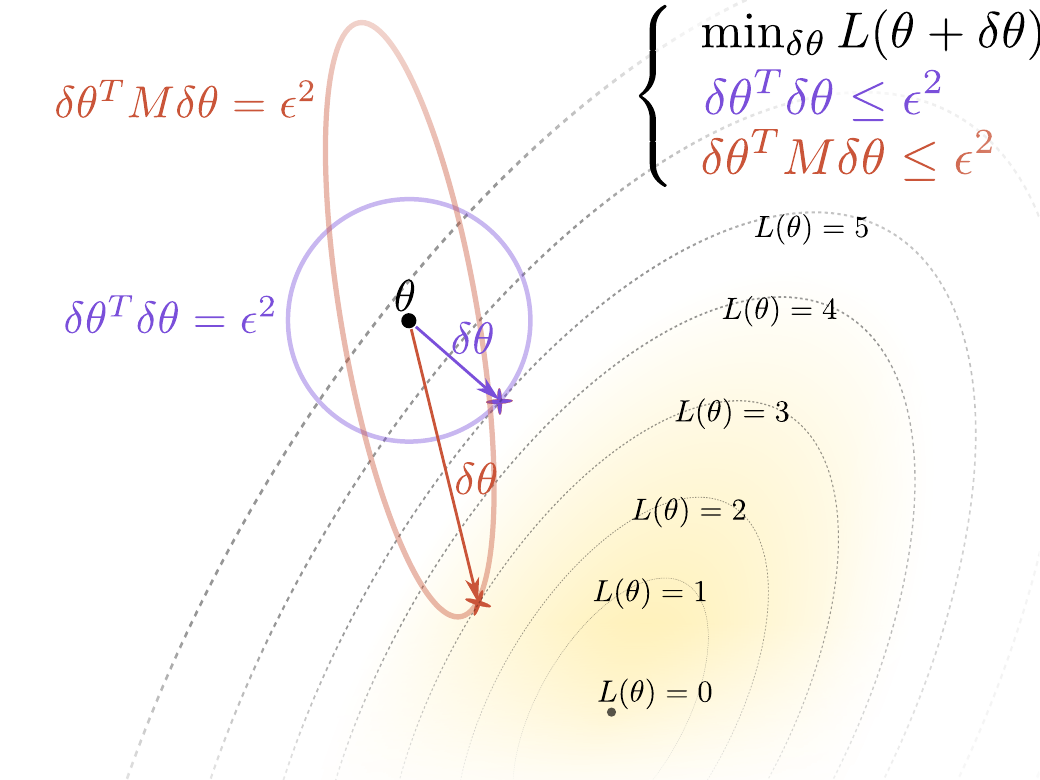}
    \caption{Different metrics affect both the gradient step size and direction.
    Here, a different $\delth$ is obtained with $M=I$ or $M$ an arbitrary symmetric positive-definite matrix.}
	\label{fig:metric_effect}
\end{figure}

Both types of approaches enhance the vanilla gradient descent update, multiplying it by the inverse of a large matrix (of size $d^2$, where $d$ is the dimensionality of the parameter space). 
We propose a simple framework that unifies these first-order or second-order improvements of the gradient descent, 
and use it to study precisely the similarities and differences between the 6 aforementioned methods.
This general framework uses a first-order approximation of the loss and constrains the step with a quadratic norm. Therefore, 
each modification $\delth$ of the vector of parameters $\vth$ is computed via an optimization problem of the following form:
\begin{equation}
  \label{eq:framework}
  \left\{
  \begin{array}{l}
    \min_{\delth} \nabla_{\vth} L(\vth)^T \delth\\ 
    \delth^T M(\vth) \delth \leq \epsilon^2,
  \end{array}
  \right.
\end{equation}
where $\nabla_{\vth} L(\vth)$ is the gradient of the loss $L(\vth)$, and $M(\vth)$ a symmetric positive-definite matrix.
The 6 methods differ by the matrix $M(\vth)$, which has an effect not only on the size of the steps, but also on the direction of the steps,
as illustrated in Figure~\ref{fig:metric_effect}.

The solution of the minimization problem~\eqref{eq:framework} has the following form (see Appendix~\ref{ap:solution}): 
$$\delth = -\alpha M(\vth)^{-1} \nabla_{\vth} L(\vth).$$

In Section~\ref{sec:first}, we show how the vanilla gradient descent, the classical Gauss-Newton method and the 
natural gradient descent method fit into the proposed framework. It can be noted that these 3 approaches constrain the steps in a way that 
is independent from the loss function.
In Section~\ref{sec:second}, we consider approaches that depend on the loss, namely the gradient covariance matrix approach, Newton's method 
and the generalized Gauss-Newton method, and show that they also fit into the framework.
Table~\ref{summary} summarizes the different values of $M(\vth)$ for all 6 approaches. 

\begin{table}
\begin{center}
\begin{tabular}{c|l}
 $M(\vth)$ & Corresponding approach\\
  \hline
  & \\
 $I$ & vanilla gradient descent \\[2pt]
 $\Esp_{\s}{\left[J_{\vx}(\vth)^TJ_{\vx}(\vth)\right]} + \lambda I$ & classical Gauss-Newton \\[4pt]
 $\Esp_\s\left[ \grad \log( p_\vth(\vy|\vx) ) \grad\log( p_\vth(\vy|\vx) )^T \right] + \lambda I$ & natural gradient (with empirical Fisher matrix) \\[4pt]
 $\Esp_{\s}\left[ \nabla_{\vth} l_\vth(\s) \nabla_{\vth} l_\vth(\s)^T \right] + \lambda I$ & gradient covariance matrix \\[4pt]
 $H(\vth) + \lambda I$ & Newton's method\\[2pt]
 $\Esp_{\s}\left[ J_{\vx}(\vth)^T \mathcal{H}_{\vy}(\hth(\vx)) J_{\vx}(\vth) \right] + \lambda I$ & generalized Gauss-Newton\\[2pt]
 \hline
\end{tabular}
\end{center}
 \caption{\label{summary} The matrices $M(\vth)$ associated to 6 popular variants of the gradient descent, when interpreted
 as instances of the optimization problem~\eqref{eq:framework}.
 See Section~\ref{sec:not} for the definitions of the notations.}
\end{table}

Although the relationships and formulations presented in this paper are known and can be found in the literature, we hope that by providing a single compact unifying view for both the first-order and second-order methods, some of the connections and differences between the 6 approaches are made more apparent. We also believe that the presentation in a common framework can facilitate the selection between these methods given a specific problem.

\section{Problem statement and notations}
\label{sec:not}

\begin{tabular}{c|l}
 Notation & Description\\
 \hline
  & \\
 $\s = (\vx, \vy)$ & a sample, with an input variable $\vx$, and an output variable $\vy$\\[2pt]
 %$d$ & dimension of the samples: $\s \in \reals^d$\\[2pt]
 $L(\vth)$ & the scalar loss to minimize, $\vth$ being the vector of parameters\\[2pt]
 $p_\vth(\cdot|\vx)$ & p.d.f. estimating the conditional distribution of the output variable\\[2pt]
 $\hth(\vx)$ & a finite-dimensional representation of the distribution $p_\vth(\cdot|\vx)$\\[2pt]
  $J_{\vx}(\vth)$ & Jacobian of the function $\vth \mapsto \hth(\vx)$\\[2pt]
 $\Esp_{\s}[\cdot]$ & expectation over the samples\\[2pt]
 $l_\vth(\s) = l(\vy, \hth(\vx))$ & the atomic loss of which $L$ is the average over the samples: $L(\vth) = \Esp_{\s}{\left[ l_\vth(\s) \right]}$\\[2pt]
 $\delth$ & small update of $\vth$ computed at every iteration \\[2pt]
 $(\cdot)^T$ & transpose operator \\[2pt]
 $\| \vec v \|$ & Euclidean norm of the vector $\vec v$: $\| \vec v \| = \sqrt{\vec v^T \vec v}$ \\[2pt]
 $\Esp_{a \sim p_\vth(\cdot|\vx)}[f(a)]$ & expected value of $f(a)$, when $a$ follows the distribution  $p_\vth(\cdot|\vx)$\\[2pt]
 $\nabla_\vth f(\vth)$ & gradient w.r.t. $\vth$: $\nabla_\vth f(\vth) = \frac{\partial f(\vth)}{\partial \vth}$ \\[2pt]
  $M_{CGN}(\vth)$ & Classical Gauss-Newton matrix: $
M_{CGN}(\vth) = \Esp_{\s}{\left[J_{\vx}(\vth)^TJ_{\vx}(\vth)\right]}
$\\[2pt]
 $\mathcal{I}_\vx(\vth)$ & Fisher information matrix of $p_\vth(\cdot|\vx)$\\[2pt]
 $F(\vth)$ & empirical Fisher matrix: $ F(\vth) = \Esp_\s\left[ \grad \log{\left( p_\vth(\vy|\vx) \right)} \grad\log{\left( p_\vth(\vy|\vx) \right)}^T \right]$\\[2pt] 
 $KL\big( p_1, p_2 \big)$ & Kullback-Leibler divergence between the probability distributions $p_1$ and $p_2$\\[2pt]
 $H(\vth)$ & Hessian of the loss $L$, defined by $\left[H(\vth)\right]_{i,j} = \frac{\partial^2 L}{\partial \vth_i \partial \vth_j}(\vth)$\\[2pt]
 $\mathcal{H}_{\vy}(\hth(\vx))$ & Hessian of the function $\vec h \mapsto l(\vy, \vec h)$ at $\vec h = \hth(\vx)$\\[2pt]
 \hline
\end{tabular}

We consider a context of regression analysis in which, based on samples $\vs = (\vx, \vy)$, the objective is to estimate the conditional distribution of $\vy$ given $\vx$.
More formally, this conditional distribution is estimated by a parametrized 
probability density function (p.d.f.) $p_\vth(\vy | \vx)$, and the goal of the learning is to progressively optimize the vector $\vth$ to improve the accuracy of this probability estimation. We furthermore assume that the 
p.d.f. $p_\vth( \cdot | \vx)$ can be represented by a finite-dimensional vector $\hth(\vx)$. For instance, in many applications, $p_\vth(\cdot | \vx)$ is
a multivariate Gaussian distribution, and in this case the vector $\hth(\vx)$ would typically contain the mean and covariance matrix components.

The accuracy of $p_\vth( \cdot | \vx)$ is measured via a loss function 
$L$ estimated over a dataset of samples $\s$. $L$ depends only on $\theta$ and we assume that it is 
expressed as the expected value (over the sample distribution) of an atomic loss, a loss \emph{per sample} $l_\vth(\s)$:
$$
L(\vth) = \Esp_{\s}{\left[ l_\vth(s) \right]}.
$$
In practice, the expected value over the samples is estimated with an empirical mean over a batch
$\mathcal{B} = (\s_1, \s_2, \dots, \s_N)$, so the loss actually used can be written 
$\hat{L}_\mathcal{B}(\vth) = \frac{1}{N}\sum_i{ l_\vth(\s_i) }$, the gradient of 
which being directly expressible from the gradient of the atomic loss (w.r.t. $\vth$). In the remainder of the paper, 
we keep expressions based on the expected value $ \Esp_{\s}{\left[ \cdot \right]}$, knowing that
at every iteration it is replaced by an empirical mean over a (new) batch of samples.

The dependency of the atomic loss to $\vth$ is via $p_\vth( \cdot | \vx)$, so we can also express it as a function of the finite-dimensional representation $\hth(\vx)$:
$$
l_\vth(\s) = l(\vy, \hth(\vx)).
$$

Remark: choosing a concrete context of regression analysis helps 
us to simplify the notations, and give examples, but the results obtained are not specific to this setting, as the 6 gradient descent variants considered can also be useful for other types of learning tasks, leading to similar expressions that can be brought back to our general framework.  

%%%%%%%%%%%%%%%%%%%%%%%%%%%%%%%%%%%%%%%%%%%%%%%%%%%%%%%%%%%%%%%%%
\section{Vanilla, classical Gauss-Newton and natural gradient descent}
\label{sec:first}

%To avoid clutter, we omit the dependence with $x$ when it is unnecessary. \\
All variants of the gradient descent are iterative numerical optimization methods: they start 
with a random initial \vth and attempt to decrease the value of $L(\vth)$ over iterations by adding a 
small increment vector \delth to \vth at each step. The core of all these algorithms 
is to determine the direction and magnitude of \delth.

\subsection{Vanilla gradient descent}
\label{sec:vanilla}
The so-called ``vanilla'' gradient descent is a first-order method that relies on a first-order Taylor approximation of the loss function $L$:
\begin{equation}
\label{eq:loss_taylor_first_order}
L(\vth + \delth) \simeq L(\vth) + \nabla_{\vth} L(\vth)^T \delth.
\end{equation}

At each iteration, the objective is the minimization of $\nabla_{\vth} L(\vth)^T \delth$ with the variable $\delth$.
If the gradient is non-zero, the value of this term is unbounded below: it suffices for instance
to set $\delth = -\alpha \nabla_{\vth} L(\vth)$ with $\alpha$ arbitrarily large.
As a result, constraints are needed to 
avoid making excessively large steps. 
In vanilla approaches, the Euclidean metric ($\|\delth\| = \sqrt{\delth^T \delth}$) is used to bound the increments $\delth$. 
The optimization problem solved at every step of the scheme is: 
\begin{equation}
  \label{eq:vanilla}
  \left\{
  \begin{array}{l}
    \min_{\delth} \nabla_{\vth} L(\vth)^T \delth \\ 
    \delth^T \delth \leq \epsilon^2,
  \end{array}
  \right.
\end{equation}
\noindent where $\epsilon$ is a user-defined upper bound. It is 
the most trivial instance of the general framework~(\ref{eq:framework}). As 
shown in Appendix~\ref{ap:solution}, the solution
of this problem is $\delth = -\alpha \grad L(\vth)$, with $\alpha = \frac{\epsilon}{\|\nabla_{\vth} L(\vth)\|}$.

To set the size of the step, instead of tuning $\epsilon$, the most common approach is 
to use the expression $-\alpha \grad L(\vth)$ and directly tune $\alpha$, which is called the \emph{learning rate}.
An interesting property with this approach is that, as $\vth$ gets closer to an optimum, the norm of the gradient $\|\nabla_{\vth} L(\vth)\|$ decreases, so 
the $\epsilon$ corresponding to the fixed $\alpha$ decreases as well.
This means that the steps tend to become smaller and smaller, which is a 
necessary property to make asymptotic convergence possible.

%%%%%%%%%%%%%%%%%%%%%%%%%%%%%%%%%%%%%%%%%%%%%%%%%%%%%%%%%%%%%%%%%
\subsection{Classical Gauss-Newton}
\label{sec:cgn}

As mentioned in Section~\ref{sec:not}, the atomic loss function $l_\vth(\s) = l(\vy, \hth(\vx))$ depends indirectly on the parameters $\vth$ via the vector $\hth(\vx)$, which is a finite-dimensional representation of the p.d.f. $p_\vth( \cdot | \vx)$. So it may be more meaningful to measure the modifications arising from an update $\delth$ by looking at the change on $\hth(\vx)$, not simply on $\delth$ as with the vanilla gradient descent approach.
The constraint $\delth^T \delth \leq \epsilon^2$ acts as if all components of $\delth$ 
had the same importance, which is not necessarily the case. Some components of $\vth$ might have much smaller effects on $\hth(\vx)$ 
than others, and this will not be taken into account with the vanilla gradient descent method, which typically performs badly with unbalanced parametrizations.
Measuring and bounding the change on the vector $\hth(\vx)$ makes the updates independent from the way $\hth$ is parametrized. To do this, a natural choice is to bound the expected squared Euclidean distance between $\hth(\vx)$ and $\hdth(\vx)$:
$$
\Esp_{\s}{\left[\|\hdth(\vx) - \hth(\vx)\|^2\right]}\leq \epsilon^2.
$$
Using again a first-order approximation, we have 
$\hdth(\vx) - \hth(\vx) \simeq J_{\vx}(\vth)\delth$, 
where $J_{\vx}(\vth)$ is the Jacobian of the function $\vth \mapsto \hth(\vx)$.
The constraint can be rewritten:
$$
\Esp_{\s}{\left[\|J_{\vx}(\vth)\delth\|^2\right]} = \delth^T\Esp_{\s}{\left[J_{\vx}(\vth)^TJ_{\vx}(\vth)\right]}\delth \leq \epsilon^2,
$$
resulting in the optimization problem:
\begin{equation}
  \label{eq:jacobian}
  \left\{
  \begin{array}{l}
    \min_{\delth} \nabla_{\vth} L(\vth)^T \delth\\ 
    \delth^T \Esp_{\s}{\left[J_{\vx}(\vth)^TJ_{\vx}(\vth)\right]} \delth \leq \epsilon^2,
  \end{array}
  \right.
\end{equation}
which fits into the general framework~\eqref{eq:framework} if the matrix $M_{CGN}(\vth) = \Esp_{\s}{\left[J_{\vx}(\vth)^TJ_{\vx}(\vth)\right]}$ is symmetric positive-definite. 

\paragraph{Damping.}
The structure of the matrix $M_{CGN}(\vth)$ makes it symmetric and 
positive semi-definite, but not necessarily definite-positive. To ensure the definite-positiveness, 
a regularization or damping term $\lambda I$ can be added, resulting
in the constraint $\delth^T \big(M_{CGN}(\vth) + \lambda I \big) \delth \leq \epsilon^2$, which can be
rewritten: 
$$
\delth^T M_{CGN}(\vth) \delth + \lambda \delth^T \delth \leq \epsilon^2.
$$
We see that this kind of damping, often called Tikhonov damping 
\citep{martens2012training}, regularizes the constraint with
a term proportional to the squared Euclidean norm of $\delth$.
It must be noted that with a regularization term, the constraint 
is not independent to the parametrization in $\vth$ anymore. And if a large value of $\lambda$ is chosen (which also usually requires
increasing $\epsilon$), the method becomes similar 
to the vanilla gradient descent.

\paragraph{A more common definition of the classical Gauss-Newton method concerns atomic losses expressed as squared errors.} 
We assume that the atomic loss $l_\vth(\s)$ is defined as follows: 
$$l_\vth(\s) = \frac{1}{2}\|\Dth(\s)\|^2,$$
where $\Delta_\vth(\s)$ is a vector-valued function.
Functions of the form $\Delta_\vth(\s) = \vy - f_\vth(\vx)$ are typical examples in the context of regression.

Denoting by $\mathcal{J}^\Delta_{\s}(\vth)$ the Jacobian of $\vth \mapsto \Dth(\s)$: 
$\mathcal{J}^\Delta_{\s}(\vth) = \left(\frac{\partial \Delta_{\vth_i}(\s)}{\partial \vth_j}\right)_{i,j}$, we have:
\begin{equation*}
\begin{split}
l_{\vth + \delta\vth}(\s) & = \frac{1}{2}
\left( \Delta_\vth(\s) + \mathcal{J}^\Delta_{\s}(\vth) \delth + O(\delth^2) \right)^T  \left( \Delta_\vth(\s) + \mathcal{J}^\Delta_{\s}(\vth)\delth + O(\delth^2) \right),
\end{split}
\end{equation*}
which can be approximated as follows by dropping part of the second-order terms: 
\\
\begin{equation*}
l_{\vth + \delta\vth}(\s) \approx l_\vth(\s) + \Delta_\vth(\s)^T \mathcal{J}^\Delta_{\s}(\vth)\delth + \frac{1}{2}\delth^T \mathcal{J}^\Delta_{\s}(\vth)^T \mathcal{J}^\Delta_{\s}(\vth)\delth
\end{equation*}
$\mathcal{J}^\Delta_{\s}(\vth)^T\Delta_\vth(\s)$ is the gradient of the loss $l$ in $\vth$, so the equation can be rewritten:
$$
l_{\vth + \delta\vth}(\s) \approx l_\vth(\s) + \grad l_\vth(\s)^T \delth + \frac{1}{2}\delth^T \mathcal{J}^\Delta_{\s}(\vth)^T \mathcal{J}^\Delta_{\s}(\vth)\delth.
$$
By averaging over the samples, we get:
$$
L(\vth + \delth) \approx L(\vth) +  \grad L(\vth)^T\delth + \frac{1}{2}\delth^T \Esp_{\s}{\left[ \mathcal{J}^\Delta_{\s}(\vth)^T \mathcal{J}^\Delta_{\s}(\vth) \right]} \delth.
$$
The classical Gauss-Newton method is often motivated by the minimization of this second-order approximation, (see \cite{bottou2018optimization}). 
Assuming that $\Esp_{\s}{\left[ \mathcal{J}^\Delta_{\s}(\vth)^T \mathcal{J}^\Delta_{\s}(\vth) \right]}$ is positive-definite, 
as shown in Appendix~\ref{ap:quad} the minimum is reached with
$$
\delth = -\Esp_{\s}{\left[ \mathcal{J}^\Delta_{\s}(\vth)^T \mathcal{J}^\Delta_{\s}(\vth) \right]}^{-1} \grad L(\vth).
$$
As in the update obtained with the optimization problem~(\ref{eq:jacobian}), the matrix with a structure of type Jacobian transpose-times-Jacobian is characteristic of the classical Gauss-Newton approach. To ensure positive-definiteness, damping can be added in the exact same way. 
The derivation that lead to~(\ref{eq:jacobian}) shows that this kind of update does not only make sense with a squared error-type of loss, so in some sense it is 
a generalization of the context in which a classical Gauss-Newton approach may be useful. 
If the dependency of the loss to $\vth$ is naturally expressed via a finite-dimensional vector $\vec v(\vth)$ (e.g. $\hth(\vx)$ or $\Dth(\s)$ in the above cases), then 
measuring the quantity $\|\vec v(\vth + \delth) - \vec v(\vth)\|$ to evaluate the magnitude of the modifications induced by $\delth$ is likely to be more meaningful than using the vanilla approach (i.e. simply measuring $\|(\vth + \delth) - \vth\| = \|\delth \| $).

\paragraph{Learning rate.}
As shown in Appendix~\ref{ap:solution}, the general framework~\eqref{eq:framework}
has a unique solution $\delth=-\alpha M(\vth)^{-1} \grad L(\vth)$, 
with $\alpha = \frac{\epsilon}{\sqrt{\grad L(\vth)^T M(\vth)^{-1}\grad L(\vth)}}$. The classical Gauss-Newton approach corresponds to $M(\vth) = M_{CGN}(\vth) + \lambda I$, or $M(\vth) = M_{CGN}(\vth)$ if we ignore the damping.
With the approach based on the second-order approximation of the loss expressed as a squared error, the resulting update has a form $\delth=- M(\vth)^{-1} \grad L(\vth)$,
which is similar to the above expression except that $\alpha = 1$. However, this theoretical difference in $\alpha$ (referred to as the learning rate in Section~\ref{sec:vanilla}) is not very significant in practice since its value is usually redefined separately. It is very common to use heuristics to set $\alpha$ to smaller values so as to increase the stability of the iterative method. Indeed,
in the proposed framework, if $\epsilon$ is constant, the updates are not getting smaller and smaller, which means that no convergence is possible.
Another example of motivation for the redefinition of $\alpha$ is that when $M(\vth)$ is a very large matrix, $M(\vth)^{-1}$ is often estimated via drastic approximations. In that case, it can be preferable to only keep the update direction of the solution of~\eqref{eq:framework}, and then perform a line search to find a value of $\alpha$ for which it is reverified that the corresponding step size is reasonable.
This line search is an important component of the popular reinforcement learning
algorithm TRPO \citep{schulman2015trust}.

\subsection{Natural gradient}
\label{sec:nat}

\begin{figure}[H]
\includegraphics[width=\textwidth]{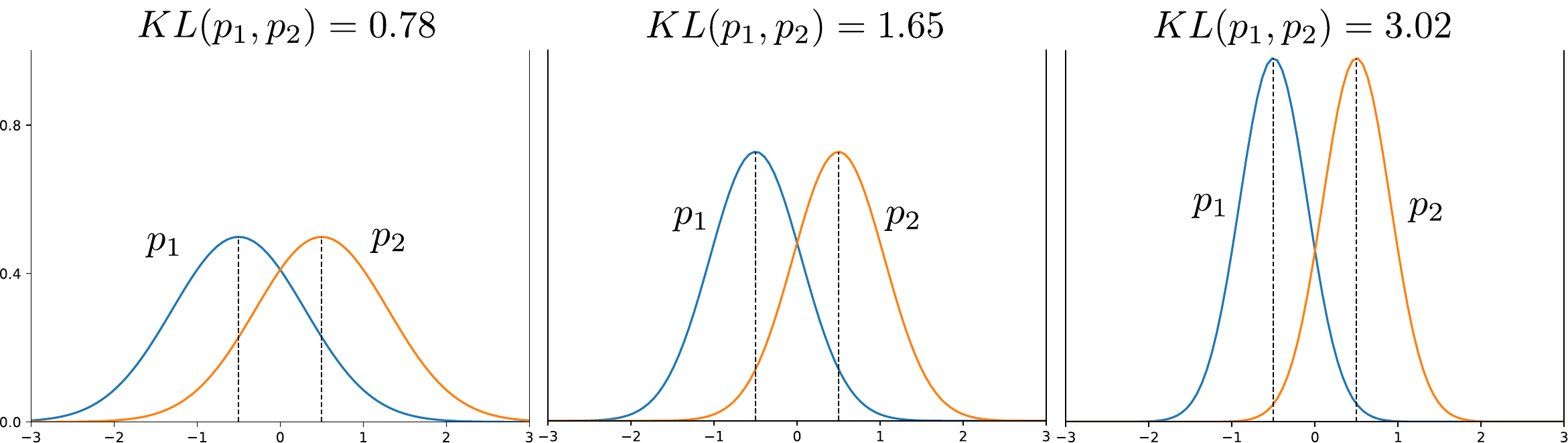}
    \caption{The same parameter change (here, a constant shift of the mean to the right) yields a larger Kullback-Leibler divergence when the variance is small.}
	\label{fig:kl}
	\end{figure}

To go one step further in terms of independence to the parametrization, it is possible to measure directly the
change from $p_\vth( \cdot | \vx)$ to $p_{\vth + \delth}( \cdot | \vx)$ with a metric on probability density functions.
This way, the updates do not even depend on the choice of
finite-dimensional representation via $\hth$.
Amari \citep{amari1997neural, amari98natural} proposed and 
 popularized the notion of natural gradient, which is based on a matrix called the 
Fisher information matrix, defined for the p.d.f $p_\vth( \cdot | \vx)$ by: 
$$
\mathcal{I}_\vx(\vth) = \Esp_{a\sim p_\vth(\cdot|\vx)}{\left[ \grad \log{\left( p_\vth(a|\vx) \right)} \grad\log{\left( p_\vth(a|\vx) \right)}^T \right]}.
$$
It can be used to measure a ``distance'' $d\ell$ between two infinitesimally close 
probability distributions $ p_\vth(\cdot|\vx)$ and $p_{\vth+\delth}(\cdot|\vx)$ as follows:
$$
d\ell^2(p_\vth(\cdot|\vx), p_{\vth+\delth}(\cdot|\vx)) = \delth^T \mathcal{I}_\vx(\vth) \delth.
$$
Averaging over the samples, we extrapolate a measure of distance between $\vth$ and $\vth + \delth$:
$$
DL^2(\vth, \vth + \delth) = \delth^T \Esp_\s\left[ \mathcal{I}_\vx(\vth) \right] \delth,
$$
where $\Esp_\s\left[ \mathcal{I}_\vx(\vth) \right] = \Esp_\s\left[ \Esp_{a\sim p_\vth(\cdot|\vx)}{\left[ \grad \log{\left( p_\vth(a|\vx) \right)} \grad\log{\left( p_\vth(a|\vx) \right)}^T \right]} \right]$ is 
the averaged Fisher information matrix.
It is common to approximate $\Esp_\s\left[ \Esp_{a \sim p_\vth(\cdot|\vx)}[\cdot] \right]$ with the 
empirical mean over the samples,
which reduces the above expression to
$$
DL^2(\vth, \vth + \delth) \approx \delth^T \Esp_\s\left[ \grad \log{\left( p_\vth(\vy|\vx) \right)} \grad\log{\left( p_\vth(\vy|\vx) \right)}^T \right] \delth.
$$
$ \Esp_\s\left[ \grad \log{\left( p_\vth(\vy|\vx) \right)} \grad\log{\left( p_\vth(\vy|\vx) \right)}^T \right]$ is called the \emph{empirical}
Fisher matrix \citep{martens2014new}. We denote it by $F(\vth)$.
Putting an upper bound on $\delth^T F(\vth) \delth$ results in the following optimization problem:
\begin{equation}
  \label{eq:natural}
  \left\{
  \begin{array}{l}
    \min_{\delth} \nabla_{\vth} L(\vth)^T \delth\\ 
    \delth^T F(\vth) \delth \leq \epsilon^2,
  \end{array}
  \right.
\end{equation}
which yields \emph{natural} gradient steps of the form
$$
\delth = -\alpha F(\vth)^{-1} \grad L(\vth),
$$
provided that $F(\vth)$ is invertible. $F(\vth)$ is always positive semi-definite. 
Therefore, as in Section~\ref{sec:cgn} with the classical Gauss-Newton approach, a damping term
can be added to ensure invertibility (but again, by doing so the independence to the parametrization is lost).
The Fisher information matrix is in some sense uniquely defined by the property of
invariance to reparametrization of the metric it induces \citep{cencov1982}, and it can be obtained 
from many different derivations. But a particularly interesting fact is that 
$d\ell^2(p_\vth(\cdot|\vx), p_{\vth+\delth}(\cdot|\vx))$ corresponds 
to the second-order approximation of the Kullback-Leibler divergence $KL(p_\vth(\cdot|\vx),p_{\vth+\delth}(\cdot|\vx))$
\citep{kullback97, akimoto2013objective}.
Hence, the terms $\delth^T \mathcal{I}_\vx(\vth) \delth$ and $\delth^T F(\vth) \delth$ share some of the properties of the Kullback-Leibler divergence. For instance, when the variance of the probability distribution $p_\vth(\cdot|\vx)$ decreases, 
the same parameter modification $\delth$ tends to result in increasingly large 
measures $\delth^T \mathcal{I}_\vx(\vth) \delth$ (see Figure~\ref{fig:kl}). 

Consequently, if the bound $\epsilon^2$ of Equation~(\ref{eq:natural}) is kept constant, the possible modifications of \vth become somehow smaller 
when the variance of the parametrized distribution decreases. Thus the natural gradient iterations slow down when the variance becomes small, which is a desirable 
property when keeping some amount of variability is important.
Typically, in the context of reinforcement learning, this variability can be related to exploration, and it should not vanish early.
This is one of the reasons why several reinforcement learning algorithms benefit from the use of natural gradient 
steps \citep{PetersNAC,schulman2015trust,wu2017scalable}.

\paragraph{Relation between natural gradient and classical Gauss-Newton approaches.}
Let us consider a very simple case where 
$p_\vth(\cdot|\vx)$ is a multivariate normal distribution with fixed covariance matrix $\Sigma = \beta^2 I$.
The only variable parameter on which the distribution $p_\vth(\cdot|\vx)$ depends is its mean $\vec \mu_\vth$, so we can use it as
representation of the distribution itself and write
$$
\hth(\vx) = \vec \mu_\vth.
$$
It can be shown that the Kullback-Leibler divergence between two normal distributions
of equal variance and different means is proportional 
to the squared Euclidean distance between the means.
More precisely, the Kullback-Leibler divergence between 
$p_\vth(\cdot|\vx)$ and $p_{\vth+\delth}(\cdot|\vx)$ is equal to 
$\frac{1}{2\beta^2}\| \hdth(\vx) - \hth(\vx)\|^2$.
For small values of $\delth$, this expression is
approximately equal to the measure 
obtained with the true Fisher information
matrix: 
$$
\frac{1}{2\beta^2}\| \hdth(\vx) - \hth(\vx)\|^2 \approx \delth^T \mathcal{I}_\vx(\vth) \delth.
$$

Bounding the average over the samples of the right term is the motivation of the natural gradient descent method.
Besides, we have seen in Section~\ref{sec:cgn} that the
classical Gauss-Newton method can be considered as a way to bound 
$\Esp_{\s}[\| \hdth(\s) - \hth(\vx)\|^2]$, which is equal to the average of the left term over the samples,
up to a multiplicative constant.
Hence, even though both methods introduce slightly different approximations,
we can conclude that in this context, 
the classical Gauss-Newton and natural gradient descent methods are very similar.
This property is used in \cite{pascanu2013revisiting} to perform a
natural gradient descent on deterministic neural networks, by interpreting
their outputs as the mean of a conditional Gaussian distribution with fixed variance.

\section{Gradient covariance matrix, Newton's method and generalized Gauss-Newton}
\label{sec:second}

The approaches seen in Section~\ref{sec:first} all fit the general framework~\eqref{eq:framework}:
\begin{equation*}
  \left\{
  \begin{array}{l}
    \min_{\delth} \nabla_{\vth} L(\vth)^T \delth\\ 
    \delth^T M(\vth) \delth \leq \epsilon^2,
  \end{array}
  \right.
\end{equation*}
with matrices $M(\vth)$ that do not depend on the loss function. But since
the loss is typically based on quantities that are relevant for the task to achieve,
it can be a good idea to exploit it to constrain the steps.
We present 3 approaches that fit into the same framework but with matrices $M(\vth)$ 
that do depend on the loss, namely the gradient covariance matrix method, 
Newton's method, and the generalized Gauss-Newton method.

\subsection{Gradient covariance matrix}

The simplest way to use the loss to measure the
magnitude of a change due to parameter modifications
is to consider the expected squared difference between 
$l_\vth(\s)$ and $l_{\vth+\delth}(\s)$:
$$
\Esp_{\s}\left[ \big(l_{\vth+\delth}(\s) - l_\vth(\s)\big)^2 \right].
$$
For a single sample \s, changing slightly the object $\vth$ 
does not necessarily modify the atomic loss $l_\vth(\s)$, 
but in many cases it can be assumed that this loss becomes 
different for at least some of the samples, yielding a positive 
value for  
$
\Esp_{\s}\left[ \big(l_{\vth+\delth}(\s) - l_\vth(\s)\big)^2 \right]
$ which quantifies in some sense the amount of change introduced by $\delth$ with 
respect to the objective. It is often a meaningful measure as it usually depends
on the most relevant features for the task to achieve.
Let us replace $l_{\vth+\delth}(\s)$ by a first-order approximation: 
$$ l_{\vth+\delth}(\s) \simeq l_\vth(\s) + \nabla_{\vth} l_\vth(\s)^T \delth.$$
The above expectation simplifies to
$$
\Esp_{\s}\left[ \big(\nabla_{\vth} l_\vth(\s)^T \delth\big)^2 \right]
= \delth^T \Esp_{\s}\left[ \nabla_{\vth} l_\vth(\s) \nabla_{\vth} l_\vth(\s)^T \right] \delth.
$$
$\Esp_{\s}\left[ \nabla_{\vth} l_\vth(\s) \nabla_{\vth} l_\vth(\s)^T \right]$ is called the 
gradient covariance matrix \citep{bottou2008tradeoffs}.
It can also be called the outer product metric~\citep{ollivier:hal-00857982}.
Putting a bound on $\delth^T \Esp_{\s}\left[ \nabla_{\vth} l_\vth(\s) \nabla_{\vth} l_\vth(\s)^T \right] \delth$, 
the iterated optimization becomes: 
\begin{equation}
  \label{eq:covgrad}
  \left\{
  \begin{array}{l}
    \min_{\delth} \nabla_{\vth} L(\vth)^T \delth \\ 
    \delth^T \Esp_{\s}\left[ \nabla_{\vth} l_\vth(\s) \nabla_{\vth} l_\vth(\s)^T \right] \delth  \leq \epsilon^2.
  \end{array}
  \right.
\end{equation}

It results in updates of the form: 
$$
\delth = -\alpha \Esp_{\s}\left[\nabla_{\vth} l_\vth(\s) \nabla_{\vth} l_\vth(\s)^T \right]^{-1} \grad L(\vth).
$$
Again, a regularization term may be added to ensure the invertibility of the matrix.
\paragraph{Link with the natural gradient.}
Let us assume that the atomic loss on a sample $\s = (\vx, \vy)$ is 
the negative log-likelihood (a very common choice):
$$
l_\vth(\s) = -\log( p_\vth(\vy|\vx) ).
$$
It follows that the empirical Fisher matrix, as defined in Section~\ref{sec:nat}, is equal 
to $\Esp_{\s}\left[ \nabla_{\vth} l_\vth(\s) \nabla_{\vth} l_\vth(\s)^T \right]$,
which is exactly the definition of the gradient covariance matrix. Therefore, in this 
case, the two approaches are identical. Several algorithms use this identity for the natural gradient
computation, e.g. \cite{george2018fast}.

\subsection{Newton's method}
\label{sec:Newton}

Let us consider now a second-order approximation of the loss: 
$$
L(\vth + \delth) \approx L(\vth) + \nabla_\vth L(\vth)^T \delth + \frac{1}{2}\delth^T H(\vth) \delth ,
$$
where $H(\vth)$ is the Hessian matrix: $\left[H(\vth)\right]_{i,j} = \frac{\partial^2 L}{\partial \vth_i \partial \vth_j}(\vth)$.
Although there are obvious counterexamples, one can argue that the first-order approximation, i.e. $L(\vth + \delth) \approx L(\vth) + \nabla_\vth L(\vth)^T \delth$ 
(which is used as minimization objective for gradient descents), is most likely good
as long as the second-order term $\frac{1}{2}\delth^T H(\vth) \delth$ remains small. Therefore, 
it makes sense to directly put an upper bound on this quantity to restrict $\delth$, as follows:
$$
\delth^T H(\vth) \delth \leq \epsilon^2.
$$
This constraint defines a trust region, i.e. a neighborhood of $\vth$
in which the first-order approximation of $L(\vth + \delth)$ is supposed to be reasonably accurate. The trust region is bounded and well defined if the matrix $H(\vth)$ is symmetric positive-definite. 
However, $H(\vth)$ is symmetric but not even necessarily positive semi-definite, unlike the matrices 
obtained with the previous 
approaches. Therefore the required damping to make it definite-positive may be larger than with other methods.
It leads to the following optimization problem solved at every iteration:
\begin{equation}
  \label{eq:hess}
  \left\{
  \begin{array}{l}
    \min_{\delth} \nabla_{\vth} L(\vth)^T \delth \\ 
    \delth^T (H(\vth) + \lambda I) \delth  \leq \epsilon^2,
  \end{array}
  \right.
\end{equation}
and to updates of the form:
$$
\delth = -\alpha (H(\vth) + \lambda I)^{-1} \grad L(\vth).
$$
\paragraph{The more usual derivation of Newton's method.} The same update 
direction is obtained by directly minimizing the damped second-order approximation:
$$
L(\vth) + \nabla_\vth L(\vth)^T \delth + \frac{1}{2}\delth^T (H(\vth) + \lambda I) \delth.
$$
When $(H(\vth) + \lambda I)$ is symmetric positive-definite, as shown in Appendix~\ref{ap:quad} the minimum of this 
expression is obtained for: 
$$
\delth = - (H(\vth) + \lambda I)^{-1} \grad L(\vth).
$$

\subsection{Generalized Gauss-Newton}

$L(\vth)$ is equal to $\Esp_{\s}{\left[ l(\vy, \hth(\vx)) \right]}$: it
does not depend directly on $\vth$ but on the outputs of $\hth$, which are vectors of finite dimension.
Posing $\delta \vec h = \hdth(\vx) - \hth(\vx)$, a second-order Taylor expansion 
of $l(\vy, \hdth(\vx))$ can be written:
$$
l(\vy, \hdth(\vx)) = l(\vy, \hth(\vx)) + \frac{\partial l(\vy, \hth(\vx))}{\partial \vec h}^T \delta \vec h
+ \frac{1}{2}\delta \vec h^T \mathcal{H}_\vy(\hth(\vx)) \delta \vec h + O(\delta \vec h^3),
$$
where $\mathcal{H}_\vy(\hth(\vx))$ is the Hessian matrix of the atomic loss $l(\vy, \hth(\vx))$
with respect to variations of $\hth(\vx)$, and $\frac{\partial l(\vy, \hth(\vx))}{\partial \vec h}$
is the gradient of $l(\vy, \hth(\vx))$ w.r.t. variations of $\hth(\vx)$.
Using the equality $\delta \vec h = J_{\vx}(\vth) \delth + O(\delth^2)$ (with $J_{\vx}(\vth)$ the Jacobian of the function $\vth \mapsto \hth(\vx)$):
\begin{equation*}
\begin{split}
l(\vy, \hdth(\vx)) = & \ l(\vy, \hth(\vx)) + \frac{\partial l(\vy, \hth(\vx))}{\partial \vec h}^T J_{\vx}(\vth) \delth
+ \frac{\partial l(\vy, \hth(\vx))}{\partial \vec h}^T O(\delth^2) \\
& + \frac{1}{2}\delth^T J_{\vx}(\vth)^T \mathcal{H}_\vy( \hth(\vx)) J_{\vx}(\vth) \delth + O(\delth^3).
\end{split}
\end{equation*}
The generalized Gauss-Newton approach is an approximation that consists in dropping
the term $\frac{\partial l(\vy, \hth(\vx))}{\partial \vec h}^T O(\delth^2)$. 
Averaging over the samples yields:
$$
L(\vth + \delth) \approx L(\vth) + \Esp_{\s}\left[\frac{\partial l(\vy, \hth(\vx))}{\partial \vec h}^T J_{\vx}(\vth)\right]^T \delth + 
\frac{1}{2}\delth^T \Esp_{\s}\left[ J_{\vx}(\vth)^T \mathcal{H}_\vy(\hth(\vx)) J_{\vx}(\vth) \right] \delth.
$$
Noticing that
$\Esp_{\s}\left[\frac{\partial l(\vy, \hth(\vx))}{\partial \vec h}^T J_{\vx}(\vth) \right] = \nabla_\vth L(\vth)$,
it results in the following approximation:
$$
L(\vth + \delth) \approx L(\vth) + \nabla_\vth L(\vth)^T \delth + 
\frac{1}{2}\delth^T \Esp_{\s}\left[ J_{\vx}(\vth)^T \mathcal{H}_\vy(\hth(\vx)) J_{\vx}(\vth) \right] \delth.
$$
As for Newton's method, the usual way to derive the generalized Gauss-Newton method 
is to directly minimize this expression (see \cite{martens2014new}), but we can
also put a bound on the quantity $\delth^T \Esp_{\s}\left[ J_{\vx}(\vth)^T \mathcal{H}_\vy(\hth(\vx)) J_{\vx}(\vth) \right] \delth$ to define a trust region for the validity of the first-order approximation (as in Section~\ref{sec:Newton}),
provided that $\Esp_{\s}\left[ J_{\vx}(\vth)^T \mathcal{H}_\vy(\hth(\vx)) J_{\vx}(\vth) \right]$ is symmetric positive-definite.
If the loss $l(\vy, \hth(\vx))$ is convex in $\hth(\vx)$ (which is often true), 
the matrix is at least positive semi-definite, so a small damping term suffices to make it positive-definite.
If a non-negligible portion of the matrices $J_{\vx}(\vth)$ are full rank,
the damping term may be added to $\mathcal{H}_\vy(\hth(\vx))$ rather than to the full matrix. 
See \cite{martens2012training} for an extensive discussion on different options for damping 
and their benefits and drawbacks. With the damping on the full matrix,
the optimization problem to solve at every iteration becomes:
\begin{equation}
  \label{eq:GGN}
  \left\{
  \begin{array}{l}
    \min_{\delth} \nabla_\vth L(\vth)^T \delth\\ 
    \delth^T \left( \Esp_{\s}\left[ J_{\vx}(\vth)^T \mathcal{H}_\vy(\hth(\vx)) J_{\vx}(\vth) \right]  + \lambda I\right) \delth \leq \epsilon^2,
  \end{array}
  \right.
\end{equation}
resulting in updates of the form: 
$$
\delth = - \alpha \left( \Esp_{\s}\left[ J_{\vx}(\vth)^T \mathcal{H}_\vy(\hth(\vx)) J_{\vx}(\vth) \right]  + \lambda I\right)^{-1} \grad L(\vth).
$$

\section{Summary and conclusion}
\label{sec:summ}

In sections~\ref{sec:first} and \ref{sec:second} we motivated and derived 6 different ways to compute parameter updates, that can all be interpreted as 
solving an optimization problem of this type:
\begin{equation*}
  \left\{
  \begin{array}{l}
    \min_{\delth} \nabla_{\vth} L(\vth)^T \delth\\ 
    \delth^T M(\vth) \delth \leq \epsilon^2,
  \end{array}
  \right.
\end{equation*}
resulting in updates of the form:
$$
\delth = - \alpha M(\vth)^{-1} \grad L(\vth),
$$

$M(\vth)$ being symmetric positive-definite. The quadratic term
of the inequality corresponds to a specific metric defined by $M(\vth)$
used to measure the magnitude of the modification induced by $\delth$. 
To evaluate this magnitude, the focus can simply be on the norm of $\delth$, 
or on the effect of $\delth$ on the loss, or on the effect
of $\delth$ on $\hth(\vx)$ or on $p_\vth(\cdot|\vx)$, resulting in various approaches,
with various definitions of $M(\vth)$. In a context of probabilistic regression, we gave 6 examples that 
correspond to popular variants of the gradient descent, summarized in Table~\ref{summary}. All methods except the natural gradient can be declined to deterministic cases.
Unifying several first-order or second-order variants of the gradient descent method
enabled us to reveal links between these different approaches,
and contexts in which some of them are equivalent.
The proposed framework gives a compact overview
of common variants of the gradient descent, and
hopefully can help
choosing adequately between them depending on the problem to solve.
Perhaps, it can also help designing new variants or combining
existing ones to obtain new desired features.

\section*{Acknowledgements}

This research was partially supported by the French National Research Agency (ANR), Project \mbox{ANR-18-CE33-0005} HUSKI.

\bibliography{local}
\bibliographystyle{iclr2018_conference}

\newpage
\appendix

\section{Solution of the optimization problem~\eqref{eq:framework}}
\label{ap:solution}

The Lagrangian of the optimization problem~\eqref{eq:framework} is 
\begin{equation*}
    \label{eq:vg_opt_lagrang}
    \mathcal{L}(\delth) = L(\vth) + \nabla_{\vth} L(\vth)^T \delth + \nu (\delth^T M(\vth) \delth - \epsilon^2),
\end{equation*}
where the scalar $\nu$ is a Lagrange multiplier. An optimal increment $\delth$ anneals the gradient of the Lagrangian w.r.t \delth,
which is equal to $\nabla_{\vth} L(\vth) + 2\nu M(\vth) \delth$. Since $M(\vth)$ is symmetric positive-definite, and therefore invertible,
the unique solution is given by $\delth = -\frac{1}{2\nu} M(\vth)^{-1} \grad L(\vth)$, which we rewrite as follows:  
  \begin{equation*}
    \label{eq:optimal_dtheta}
    \delth = -\alpha M(\vth)^{-1} \grad L(\vth).
  \end{equation*}
Plugging this expression in problem~\eqref{eq:framework} yields the following minimization problem with variable $\alpha$: 
\begin{equation*}
  \left\{
  \begin{array}{l}
    \min_{\alpha} -\alpha\nabla_{\vth} L(\vth)^T M(\vth)^{-1} \grad L(\vth)\\ 
    \alpha^2\nabla_{\vth} L(\vth)^T M(\vth)^{-1} \grad L(\vth) \leq \epsilon^2,
  \end{array}
  \right.
\end{equation*}
and assuming that the gradient $\nabla_{\vth} L(\vth)$ is non-zero, the optimum is reached for: 
\begin{equation*}
\alpha = \frac{\epsilon}{\sqrt{\grad L(\vth)^T M(\vth)^{-1}\grad L(\vth)}}.
\end{equation*}

\section{Minimization of a quadratic form}
\label{ap:quad}

Let us consider a function $f(\delth) = c + \vec g^T \delth + \frac{1}{2} \delth^T M(\vth) \delth$,
where $c$ is a scalar, $\vec g$ a vector and $M(\vth)$ a symmetric positive-definite matrix.
The gradient of $f$ is:
$$ \nabla_{\delth} f(\delth) = \vec g + M(\vth) \delth. $$
$M(\vth)$ being invertible, this gradient has a unique zero that corresponds 
to the global minimum of $f$:
$$
\delth^* = -M(\vth)^{-1}\vec g.
$$

\end{document}